%% file: main.tex
\documentclass{article}
\usepackage{arxiv}
\usepackage[noadjust]{cite}
\usepackage{graphicx}
\usepackage{tikz}

\usepackage{amsmath,amssymb}
\definecolor{cistiblightgray}{RGB}{245,246,246}
\definecolor{cistibpurple}{RGB}{78,68,120}
\definecolor{strongpurple}{RGB}{130,70,200}
\definecolor{cistibgreen}{RGB}{90,162,82}
\usepackage{algorithmic}
\usepackage{algorithm}

\begin{document}

\title{Partially Conditioned Generative Adversarial Networks}

\author{
Francisco J. Ibarrola
\thanks{\footnotesize CISTIB Centre for Computational Imaging \& Simulation Technologies in Biomedicine, School of Computing, University of Leeds, UK. ({\tt f.ibarrola@leeds.ac.uk})}
\thanks{\footnotesize
LICAMM Leeds Institute for Cardiovascular and Metabolic Medicine, School of Medicine, University of Leeds, UK.}
\And Nishant Ravikumar
$^{\ast \dagger}$
\And Alejandro F. Frangi
$^{\ast \dagger}$
\thanks{\footnotesize Department of Cardiovascular Sciences, and Department of Electrical Engineering, ESAT/PSI, KU Leuven, Leuven, Belgium; Medical Imaging Research Center, UZ Leuven, Herestraat 49, 3000 Leuven, Belgium.}
}

\maketitle
\begin{abstract}
Generative models are undoubtedly a hot topic in Artificial Intelligence, among which the most common type is Generative Adversarial Networks (GANs). These architectures let one synthesise artificial datasets by implicitly modelling the underlying probability distribution of a real-world training dataset. With the introduction of Conditional GANs and their variants, these methods were extended to generating samples conditioned on ancillary information available for each sample within the dataset. From a practical standpoint, however, one might desire to generate data conditioned on partial information. That is, only a subset of the ancillary conditioning variables might be of interest when synthesising data. In this work, we argue that standard Conditional GANs are not suitable for such a task and propose a new Adversarial Network architecture and training strategy to deal with the ensuing problems. Experiments illustrating the value of the proposed approach in digit and face image synthesis under partial conditioning information are presented, showing that the proposed method can effectively outperform the standard approach under these circumstances.
\end{abstract}
\textbf{keywords:}
Generative models, partial information, conditional GANs, image synthesis.

\section{Introduction}

Throughout the past years, much research effort has been put in generative model development (\cite{goodfellow2014generative}, \cite{kingma2013auto}). On top of building a model capable of synthesising realisations from the underlying distribution of a given data set, one might want to generate samples conditioned by certain information. While this problem has been addressed for synthesising data with certain attributes (\cite{mirza2014conditional}), some fields, such as virtual patient synthesis for medical testing (\cite{lassila2020population}, \cite{abadi2020virtual}) could greatly benefit from a model's ability to generate samples from an arbitrary subset within a large set of conditioning information. Building a model which can generate from partial conditioning information is the problem we shall address in what follows.

Let us consider the problem of generating synthetic data samples conditioned on some ancillary information. Assume we have a data set consisting of pairs $\{x_n,y_n\}_{n=1, \ldots, N}$, where $x_n\in \mathbb{R}^M$ is the target data and $y_n \in \mathbb{R}^K$ is a vector containing available information associated to $x_n$. Let us further assume that the pairs $\{x_n,y_n\}$ are independently sampled realizations from an underlying probability distribution $\pi(X,Y)$. Then the goal of a conditional generative model is, given an arbitrary entry $y\in \mathbb{R}^K$, to generate representative samples from the conditional distribution $\pi(X|Y=y)$.

A recent yet widespread approach to dealing with these problems is the use of Conditional GANs (\cite{mirza2014conditional}) where two distinct Neural Networks, a Generator and a Discriminator, are trained in an adversarial fashion. This results in a Generator producing samples from $\pi(X|Y=y)$ from random Gaussian noise, given some conditioning information $y$.

Variants of this approach have been successfully used in various applications, such as medical image synthesis (\cite{xu2019multichannel}), image deblurring (\cite{choi2020modified}) and parameter optimisation (\cite{alonso2020image}), to name a few. Nevertheless, open problems that are yet to be addressed might be encountered when dealing with real world scenarios.

A straightforward data synthesis problem might entail generating samples conditioned only on partial information rather than the complete conditioning vector $y$. Given an incomplete conditioning vector $\bar{y}$, we want to sample from $\pi(X|Y=\bar{y})$. Furthermore, the components of $y$ kept in $\bar{y}$ might be unknown \emph{a priori} or they might even change depending on the available data and formulation of the problem. For instance, in the field of medical image synthesis, there is usually plenty of conditioning information with which to train a model. From a practical standpoint, however, it is usually desirable for some characteristics (conditioning variables) to be left unspecified when generating populations.

While other authors have tackled the problem of training using databases with noisy or uncertain conditioning variables (\cite{thekumparampil2018robustness}, \cite{thekumparampil2019robust}), to the best of our knowledge the problem of building a generator that can work under partial conditioning has yet to be addressed.

In this work we show standard Conditional GANs fail in this problem setting, thus highlighting the need to either retrain the model for every set of conditioning variables or to infer the missing inputs by learning their joint latent space, which is often non-trivial. To overcome this issue, we propose an adversarial architecture that generates samples from incomplete conditioning vectors without estimating the missing entries. We then describe a simple learning strategy and illustrate and validate our approach in two classical problems in computer vision - synthesis of handwritten digit (MNIST) and face images (CelebA).

\section{Conditional GANs}

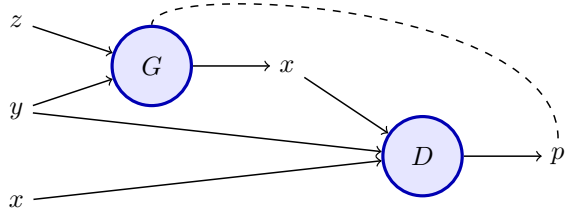
\begin{figure}
    \centering
    \input{frames/tikzfig_CGAN.tex}
    \caption{Diagram of the Conditional GAN network architecture. $G$ corresponds to the generator network and $D$ to the discriminator.}
    \label{fig:CGAN_arch}
\end{figure}

Let $\{x,y\}$ be a realisation from an unknown distribution $\pi(X,Y)$, and let us consider an additional random variable $Z \sim \mathcal{N}(0,I_{J\times J})$.  A Conditional GAN (\cite{mirza2014conditional}) consists of a generator function $G:\mathbb{R}^J \times \mathbb{R}^K \rightarrow \mathbb{R}^M$ which seeks to generate artificial realisations from $\pi(X|Y=y)$, and a discriminator function $D:\mathbb{R}^M \times \mathbb{R}^K\rightarrow [0,1]$ that estimates the probability of an input $x\in\mathbb{R}^M$ coming from the actual training set rather than the generator, given the corresponding conditioning vector $y$. Hence, training a Conditional GAN equals to solving a two-player min-max game over a cost function, given by
\begin{align}\label{eqn:Vcgan}
    V(G,D,y) \doteq & \; \mathbb{E}_{x\sim \pi_{data}}\log D(x|y) +\mathbb{E}_{z\sim \pi_z}\log [1-D(G(z|y)|y)].\nonumber
\end{align}
Given the characteristics of the function, $D$ should be chosen as to maximise $V$ (which translates into good discrimination capability), while $G$ should be such as to attempt to minimise $V$, thus ``fooling'' the discriminator. This structure is illustrated in Figure \ref{fig:CGAN_arch}.

Once this network has been trained, sampling from $\pi(X|Y=y)$ amounts to evaluating $G(\,\cdot\, |y)$ on a realisation $z$ from the distribution $\pi_z$. But if we want to generate samples conditioned on a subset $\bar{y}$ of the conditioning variables, $G(z|\bar{y})$ will generally not produce good results during inference.  

\subsection{Missing conditioning information}

When working with missing entries in the conditioning vector, one can address two issues: the first is how to represent the missing data in the context of the network inputs; the second is how to synthesise data under partial conditioning information.

Let us consider that the information on $\mathbb{Y}$ is categorical, \emph{i. e.} $y$ contains one or more characteristics describing $x$. Then, given $L$ possible states, we can define $y \in \{0,1\}^L$ as
\begin{align}
y_l = 
\begin{cases}
1 & \text{if $x$ is in the $l$-th class,} \\
0 & \text{otherwise.}
\end{cases}
\end{align}

This kind of representation is called \emph{one-hot-embedding} (\cite{harris2010digital}). Using this definition, missing information can simply be noted as an element of $y$ being $0$ where it should be $1$.

\begin{figure}
    \centering
    
    $y = (0, 0, 0, 1, 0, 0, 0, 0, 0, 0)$
    
    \includegraphics[width = 0.5\columnwidth]{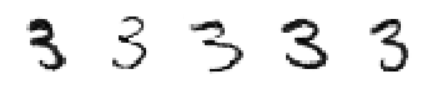}
    
    \smallskip
    
    $y = (0, 0, 0, 0, 0, 0, 0, 0, 0, 0)$
    
    \includegraphics[width = 0.5\columnwidth]{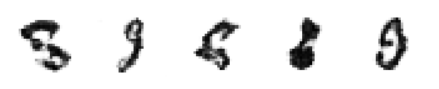}
    \caption{Examples of images generated with a standard Conditional GAN. Top row: complete conditioning information. Bottom row: incomplete conditioning information.}
    \label{fig:MINST_std}
\end{figure}

Suppose we have trained our Conditional GAN with the aforementioned vector $y$, the question remains whether the generator would still work with incomplete data. In order to illustrate the ensuing issue, we trained a Conditional GAN using the MNIST dataset \cite{lecun1998gradient}, using a vector $y \in \{0,1\}^{10}$ where a 1 in the $i$-th position indicates the $i$-th digit. An example of the generator's output is given in the top row images on Figure \ref{fig:MINST_std}. The bottom row, on the other hand, depicts what happens when the label is missing, which ideally should be a drawn digit, or at least resemble one.

To demonstrate this is a fundamental problem of this approach rather than a consequence of the complete lack of conditioning information given to the network, we have repeated the experiment using $y \in \{0,1\}^{11}$, using the label $y = (0, 0, 0, 0, 0, 0, 0, 0, 0, 1, 1)$ for indicating the digit 9 while preserving all the others.

\begin{figure}
    \centering
    
    $y = (0, 0, 0, 0, 0, 0, 0, 0, 0, 1, 1)$
    
    \includegraphics[width = 0.5\columnwidth]{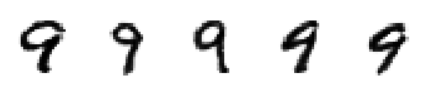}
    
    \smallskip
    
    $y = (0, 0, 0, 0, 0, 0, 0, 0, 0, 1, 0)$
    
    \includegraphics[width = 0.5\columnwidth]{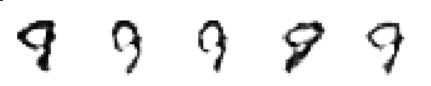}
    
    \smallskip
    
    $y = (0, 0, 0, 0, 0, 0, 0, 0, 0, 0, 1)$
    
    \includegraphics[width = 0.5\columnwidth]{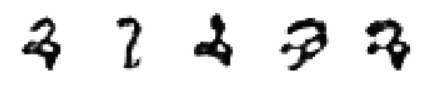}
    \caption{Output examples of a standard Conditional GAN Generator under complete conditioning information (top row) and missing conditioners (middle and bottom rows).}
    \label{fig:MINST_99}
\end{figure}

Figure \ref{fig:MINST_99} depicts the results obtained when the generator is given missing inputs, whilst preserving the necessary information for producing the desired output (digit 9). The alluded problem persists.

One might then seek to ``fill'' the missing entries on $\bar{y}$, obtaining an approximation $\hat{y}\approx y$ and then feed it to the Conditional GAN. While this could work in a toy example such as the one depicted above, when dealing with real data, ``filling'' $y$ amounts to sampling from the conditional distribution $\pi(Y|Y_l = \bar{y}_l, \,\forall l\in\mathcal{L}),$ where $\mathcal{L}$ is the set of indexes of available data. This problem can be rather simple when the random variables $Y_l$ are independent, but in more practical and interesting scenarios, there is always some degree of correlation between the conditioning variables, which implies that sampling from the conditional distribution $\pi(Y|Y_l = \bar{y}_l, \,\forall l\in\mathcal{L})$ becomes a non-trivial problem.

In the next Section we propose an architecture to tackle the missing data problem in a general context that also precludes the need of estimating the aforementioned marginal distribution.

\section{Architectures for dealing with missing conditioning data}

From what has been observed in Figure \ref{fig:MINST_99}, one can readily conclude that the network is not actually learning a 1 in the 10-th entry of $y$ and a 1 in the 11-th entry as separate characteristics indicating a digit 9. Hence, the problem in this simple amounts to making the network learn an OR gate over the 10-th and 11-th entries.

The primary characteristic with which we would like to imbue our network is the ability to utilise the available information when generating data, even if incomplete. In the problem illustrated in Figure \ref{fig:MINST_99}, this would mean learning an OR gate between the redundant information bits. In the context of a real world problem, this would mean the network be able to bypass the missing information and generate samples from whatever data is available.

Formally, given a vector $\bar{y}$ with missing entries, we want our network to be able to sample from $\pi(X|Y_l = \bar{y}_l, \,\forall l\in\mathcal{L})$. Note that accomplishing this would overcome the problem of estimating $\pi(Y|Y_l = \bar{y}_l, \,\forall l\in\mathcal{L}).$ 

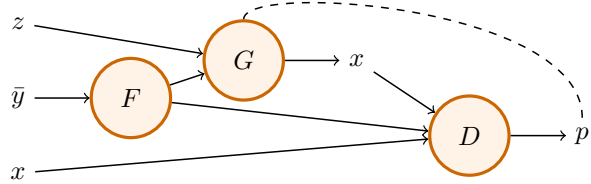
\begin{figure}
    \centering
    \input{frames/tikzfig_CGAN2.tex}
    \caption{Diagram of PCGAN architecture. $G$ corresponds to the generator, $D$ to the discriminator and $F$ to the feature extraction network.}
    \label{fig:CGAN_arch2}
\end{figure}

To do this, we propose a Partially Conditioned Generative Adversarial Network (PCGAN), formulated as shown in Figure \ref{fig:CGAN_arch2}. In this diagram, $F$ is a ``feature extraction'' neural network that extracts the underlying information from $\bar{y}$. This can be accomplished by a shallow multilayer perceptron for categorical information, but if other type of conditioning information is available, other architectures might be better suited (e.g. if the conditioner is an image, we could choose $F$ to be a convolutional network). In our toy example, the goal of $F$ would be to act as an OR gate.

This means that the cost function is now
\begin{align*}\label{eqn:Vcgan}
    V(G,D;\bar{y}) \doteq & \; \mathbb{E}_{x\sim \pi_{data}}\log D(x|F(\bar{y})) +\mathbb{E}_{z\sim \pi_z}\log [1-D(G(z|F(\bar{y}))\,|\,F(\bar{y}))].
\end{align*}

Note that since $F$ is a differentiable neural network, adapting the backpropagation process for optimisation is straightforward.

Once we have an architecture capable of information extraction (building an OR gate), the problem remains on how to embed that function into the architecture. A simple solution is that for the generator to work with missing conditioning entries, it must learn to work with missing entries. Hence, during the training process we shall remove some of the entries on the data points to emulate the expected working condition of the generator when synthesising data. A simple way to accomplish this is to assign a probability $p\in(0,1)$ of being observed for every entry. The training process is summarized in Algorithm \ref{alg:PCGAN}.

\begin{algorithm}
\caption{PCGAN training}
\label{alg:PCGAN}

\medskip
\begin{algorithmic}


\FOR{ $\{x_i,y_i\} \in \{X,Y\} $}

\medskip

\STATE \textbf{Updating $D$}
\STATE Let $z\sim \pi_z$ and $\bar{y}_i = y_i \odot \mathbf{b}_p$
\STATE $\hat{x}_i = G(z,F(\bar{y}_i))$
\STATE $e_\text{fake} = D(\hat{x}_i,F(\bar{y}_i))$
\STATE $e_\text{true} = D(x_i,F(\bar{y}_i))$
\STATE Backpropagate $e_\text{fake}+e_\text{true}$ to update the weights of $D$.

\medskip

\STATE \textbf{Updating $G$ and $F$}
\STATE Let $z\sim \pi_z$ and $\bar{y}_i = y_i \odot \mathbf{b}_p$
\STATE $\hat{x}_i = G(z,F(\bar{y}_i))$
\STATE $e_\text{fake} = D(\hat{x}_i,F(\bar{y}_i))$
\STATE Backpropagate $e_\text{fake}$ to update the weights of $G$ and $F$.

\medskip

\ENDFOR

\hrulefill
\STATE \small{ $\odot$ denotes the Hadamard (pointwise) product.}
\STATE \small{ $\mathbf{b}_p$ denotes a vector of independent realizations from a binomial distribution with parameter $p$.}
\end{algorithmic}
\end{algorithm}

\medskip

 Note that $F$ is trained along with $G$ as the goal is to tune it for the generation process. The parameter $p$ corresponds to the percentage of conditioning entries to be used for training, and does not necessarily match that to be observed when generating samples in a practical problem. Finally, while the training process is described for using a sample at a time, batch training is recommended.

The next section illustrates through examples how the proposed method works, as well as its performance.

\section{Experiments}

Three experiments will be presented. The first one for demonstrating the issue stated in the toy example in Section 2 is effectively solved. The second experiment is designed to assess performance of the method on digit synthesis with artificial conditioners, where the simplicity of the images helps make the interpretation of results easier. Finally, a third experiment on conditional face image synthesis shall help better demonstrating the performance of the approach on a real world scenario.

\subsection{Digit synthesis with duplicate labels}

\begin{figure}[ht]
    \centering
    
    $y = (0, 0, 0, 0, 0, 0, 0, 0, 0, 1, 1)$
    
    \includegraphics[width = 0.5\columnwidth]{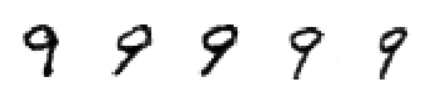}
    
    \smallskip
    
    $y = (0, 0, 0, 0, 0, 0, 0, 0, 0, 1, 0)$
    
    \includegraphics[width = 0.5\columnwidth]{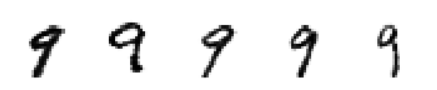}
    
    \smallskip
    
    $y = (0, 0, 0, 0, 0, 0, 0, 0, 0, 0, 1)$
    
    \includegraphics[width = 0.5\columnwidth]{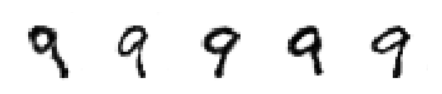}
    \caption{Output examples of a PCGAN Generator using incomplete conditioners.}
    \label{fig:PCGAN_99}
\end{figure}

In the first experiment we address the issue highlighted in Figure \ref{fig:MINST_99}. To do that, we simply rerun the experiment of doubling the conditioner for digit 9, and this time we trained the architecture depicted in Figure \ref{fig:CGAN_arch2}, with a 30\% chance of missing one of the bits corresponding to 9 during training, and defining $F$ as a single-layer perceptron. Then, we tested the network using missing information, and obtained the results depicted in Figure \ref{fig:PCGAN_99}. These results help demonstrate that the network is able to learn that either 10th OR the 11th component of the input vector correspond to the digit 9.

\subsection{Handwritten digit synthesis from partial conditioning information}

In order to perform this experiment, we built some synthetic features based upon the MNIST handwritten digits training dataset (\cite{lee2019collagan}) digit labels. We took the set $\mathbb{N}^9_0 = \{0, 1, \dots, 9\}$ and made a random partition into three sets $\{P^{(1)}_1, P^{(1)}_2, P^{(1)}_3\}$ with at least 3 elements each. Then, for every element $n\in \mathbb{N}^9_0$, we assigned the conditioner
\begin{equation*}
    y^{(1)} = \begin{cases}
    (1, 0, 0) &\text{if} \quad n\in P^{(1)}_1 \\
    (0, 1, 0) &\text{if} \quad n\in P^{(1)}_2 \\
    (0, 0, 1) &\text{if} \quad n\in P^{(1)}_3
    \end{cases}
\end{equation*}
We repeated this 10 times to produce the conditioners $y^{(1)}, y^{(2)}, \ldots, y^{(10)}$, so every element $n\in \mathbb{N}^9_0$ has an associated binary conditioning vector $y = [y^{(1)}, y^{(2)}, \ldots, y^{(10)}]$ of size 30. This procedure produces artificial conditioners which are redundant, in the sense that if $A = \cap_{k=1}^{10} P^{(k)}_{i_k}$, there exists $m:\;A = \cap_{k\neq m} P^{(k)}_{i_k}$. This means that the information can be characterised by a subset of the partitions, and hence a model well suited for the proposed problem should exhibit robustness regarding missing 1's on $y$.

\begin{figure}
\centering
    
CGAN
    
\includegraphics[width=0.5\columnwidth]{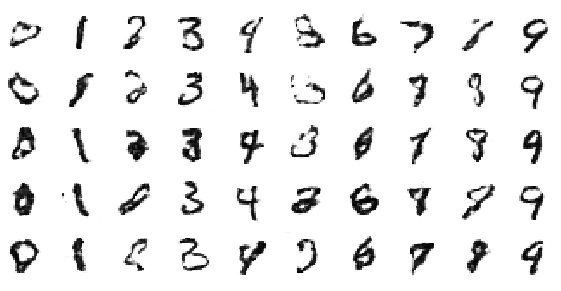} \smallskip
    
PCGAN
    
\includegraphics[width=0.5\columnwidth]{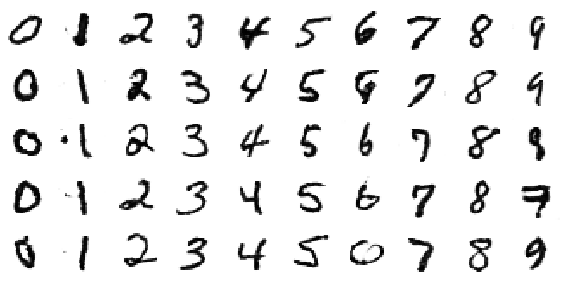}
\caption{Output examples of Generators, when conditioned on 30\% missing conditioning entries. CGAN: Standard Conditional GAN. PCGAN: Partially Conditioned GAN, trained with 15\% missing conditioning entries. }
    \label{fig:MINST_CGANvsPCGAN}
\end{figure}

Using these artificial features, we have tested our proposed network architecture (trained with missing information) against a standard Condtional GAN. We have taken the precaution to prevent the additional layer $F$ from adding capacity to the proposed architecture (see Appendix \ref{fig:CGAN_arch} for details). The results generated with these networks when testing with 30\% missing conditioners (chosen randomly) are illustrated in Figure \ref{fig:MINST_CGANvsPCGAN}. Some of the images generated by a standard Conditional GAN mix digits do not resemble any target digits (\emph{i. e.} any of the original classes). The proposed PCGAN, however, is seen to produce better results in terms of image quality.



\begin{figure}
    \centering
    \includegraphics[width = 0.45\columnwidth]{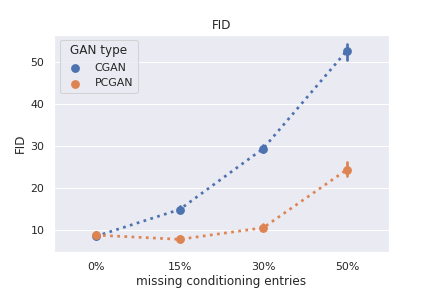}
    \includegraphics[width = 0.45\columnwidth]{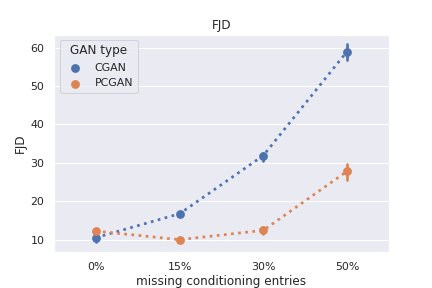}
    \caption{FID and FJD measures for digit generation. CGAN: Standard Conditional GAN. PCGAN: Partially Conditioned GAN, trained with 15\% missing conditioning entries.}
    \label{fig:FID_FJD_MINST}
\end{figure}

As a way of quantifying the differences between the methods, we made use of the Fr\'echet Inception Distance (FID, \cite{szegedy2016rethinking}), where both the real images and model samples are embedded in a learnt feature space, on which the Fr\'echet distance is computed. This performance measure, however, does not consider the conditioning information, so we also measured the Fr\'echet Joint Distance (FJD, \cite{devries2019evaluation}), which makes use of a joint image-conditioning embedding space. The performance measures obtained when generating using $0\%$, $15\%$, $30\%$ and $50\%$ missing entries are depicted in Figure \ref{fig:FID_FJD_MINST}. The numbers account for similar FID/FJD values when synthesising with a Conditional GAN or PCGAN from complete conditioning information, but consistently with what was observed on Figure \ref{fig:MINST_CGANvsPCGAN}, the PCGAN exhibits greater robustness with respect to missing conditioning entries. In fact, the difference becomes larger as more conditioners are randomly removed when synthesising.  



\subsection{Face image generation}

\begin{figure}
    \centering
    CGAN
    \smallskip
    
    \includegraphics[width = 0.5\columnwidth]{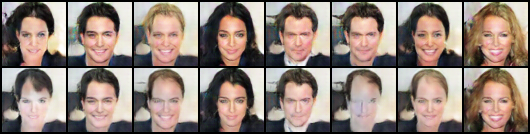}
    
    \smallskip
    
    \includegraphics[width =  0.5\columnwidth]{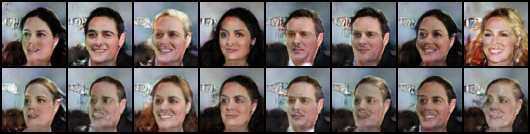}
    
    \smallskip
    
    \includegraphics[width = 0.5\columnwidth]{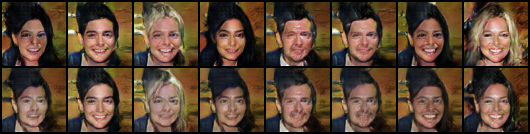}
    
    \medskip
    
    PCGAN
    \smallskip
    
    \includegraphics[width = 0.5\columnwidth]{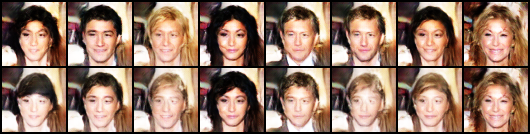}
    
    \smallskip
    
    \includegraphics[width = 0.5\columnwidth]{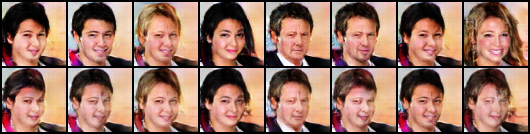}
    
    \smallskip
    
    \includegraphics[width = 0.5\columnwidth]{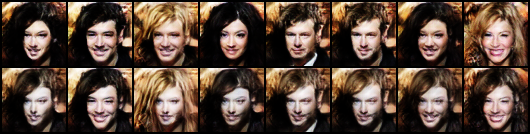}
    
    \caption{Synthetic face images generated using Conditional GAN and PCGAN. On every set of images, the top row corresponds to complete conditioning information, and bottom row has 30\% missing inputs (at random). For fairness in the comparisons, the conditioning entries and missing entries are the same for both networks.}
    \label{fig:celeba_gen_samples}
\end{figure}


\begin{figure}
    \centering
    \includegraphics[width = 0.45\columnwidth]{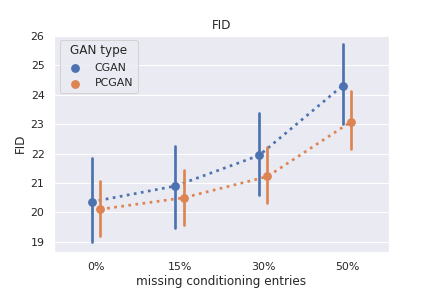}
    \includegraphics[width = 0.45\columnwidth]{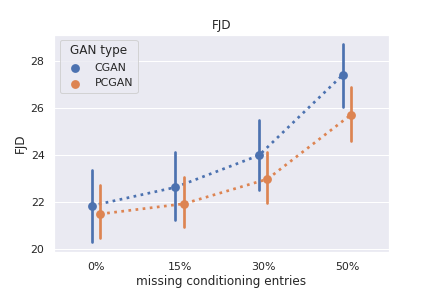}
    \caption{FID and FJD measures for face generation. CGAN: Standard Conditional GAN. PCGAN: Partially Conditioned GAN, trained with 15\% missing conditioning entries.}
    \label{fig:celeba_CGANvsPCGAN}
\end{figure}

In order to illustrate how the use of PCGANs translates into a real world application, we shall consider the problem of conditioned human face image generation. To do this, we utilise the CelebA dataset (\cite{liu2015faceattributes}), which contains human face images along with 40 binary annotations associated with visual characteristics, such as hair colour, expression and skin tone.

Once the network has been trained, if one wants to generate samples from a subset of the conditioning information, it is clear the remaining conditioning features cannot be randomly guessed. Otherwise, one might end up asking the network to synthesize faces with incompatible features, such as \emph{bald} and \emph{brown hair}. To demonstrate that PCGAN can overcome this issue, we trained and tested a Conditional GAN and a PCGAN on the CelebA data set. Results are depicted in Figure \ref{fig:celeba_gen_samples}, where three sets of synthetic images are shown for each method (architecture and model details are in Appendix \ref{sec_arCelebA}). On each set, the top row corresponds to images generated with complete conditioning information, and the bottom row shows images generated with 30\% missing entries (completely at random). In some cases, depending on the attributes, standard Conditional GANs produce samples with some degree of degradation, while the PCGAN exhibits more robustness to the same conditions. For fairness of comparison, the displayed $n-th$ sample of the Conditional GAN and the $n-th$ sample of the PCGAN have the same present and missing attributes.

Performance was evaluated again by measuring the FID and FJD. Figure \ref{fig:celeba_CGANvsPCGAN} shows the mean and standard deviation of the measures for ten different trials of each network, and for 4 different degrees of missing conditioning entries (completely at random). As in the case of the previous experiments, both networks exhibit similar performance when synthesising from the complete conditioning set of variables, but performance decays more rapidly for the standard Conditional GAN when the proportion of removed conditioning entries increases. The difference in decay in this case is milder than the one using MNIST, which is consistent with the fact that the conditioning attributes are much less correlated than in the previous experiment. 



\section{Conclusions and future work}

In this study we addressed the problem of building a generative model that can work under partial conditioning information. To do so, we proposed a network architecture that allows for generating from the features that underlay the conditioning information, as well as a strategy for properly training the network.

The experiments conducted demonstrate that the proposed approach can successfully generate images even under a significant proportion of missing conditioning information, unlike the widely used Conditional GAN architecture. Furthermore, it is immediately clear that the architecture is robust to training under some degree of missing conditioning data.

In the future, we intend to extend the proposed model to be able to work with continuous conditioning variables and to find ways to optimally choose the architecture parameters. Additionally, we envisage the use of the concepts proposed herein, for the purpose of anatomical shape synthesis, conditioned on patient meta data.

\appendix{Network structures}

In what follows, we present a description of the Network architectures used for the experiments. Should the reader want any additional details, a Python notebook to reproduce the presented experiments can be found at https://github.com/fibarrola/pcgan

\subsection{Architecture for MNIST}\label{sec_arMNIST}

\begin{itemize}
\item \emph{Parameters - } Learning rate = $1\times 10^{-4}$. Latent space dimension = 10. Batch size = 128.
\item\emph{Generator - } A fully connected layer $F$ is applied to the conditioning vector $y$, then concatenated with the noise vector $z$. Then a second fully connected layer and two transposed convolutional layers are applied. Batch normalization is used after every convolutional layer. Activation functions are sigmoid for the last layer and ReLU for all the others.
\item\emph{Discriminator for Conditional GAN - } Images go through two consecutive convolutional layers and batch normalization. A fully connected layer $F_2$ is applied to $y$. Then the outputs are flattened, concatenated, and put through two more fully connected layers. Activation functions are sigmoid for the last layer and ReLU for all the others.
\item\emph{Discriminator for PCGAN - } Same architecture as discriminator for standard Conditional GAN, but $F_2 = F$.
\end{itemize}

\subsection{Architecture for CelebA}\label{sec_arCelebA}

\begin{itemize}
\item\emph{Parameters - } Learning rate = $2\times 10^{-4}$. Latent space dimension = 100. Batch size = 64.
\item\emph{Generator - } A transposed convolutional layer $F$ is applied to the conditioning vector $y$, then concatenated with the output of putting the noise vector $z$ through another transposed convolutional layer. Four transposed convolutional layers are applied afterwards. Batch normalisation is used after every convolutional layer. Activation functions are hyperbolic tangent for the last layer and ReLU for all the others.
\item\emph{Discriminator for Conditional GAN - } A fully connected layer is applied to $y$, and the result is stacked as a 4-th image channel.  Then the outputs are fed through five convolutional layers. Activation functions are sigmoid for the last layer and Leaky ReLU (negative slope = 0.2) for all the others.
\item\emph{Discriminator for PCGAN - } The same architecture than the discriminator for the standard Conditional GAN, but it is fed $F(y)$ instead of $y$.
\end{itemize}

\section*{Acknowledgement}

AFF is supported by the RAEng Chair in Emerging Technologies (CiET1819/19).

\bibliographystyle{IEEEtran}
\bibliography{mybibfile}

\end{document}

%% file: frames/tikzfig_CGAN.tex
\begin{tikzpicture}[scale=1.2,
roundnode/.style={circle, draw=blue!70!black, fill=blue!10, very thick, minimum size=30},
squarednode/.style={rectangle, draw=cistibpurple!80, fill=cistibpurple!10, very thick, minimum size=1},line width = 0.2mm,node distance=2cm]

\node at (0, 2)   (z) {$z$};
\node at (0, 1)   (y) {$y$};
\node at (0, 0)   (x) {$x$};
\node[roundnode] at (1.5, 1.5)   (G) {$G$};
\node[roundnode] at (4.5, 0.5)   (D) {$D$};
\node at (3, 1.5)   (x2) {$x$};
\node at (6, 0.5)   (p) {$p$};

\draw[->] (z)--(G);
\draw[->] (y)--(G);
\draw[->] (y)--(D);
\draw[->] (G)--(x2);
\draw[->] (x2)--(D);
\draw[->] (x)--(D);
\draw[->] (D)--(p);
\draw[dashed] (p.north).. controls +(up:1.5) and +(up:0.5) ..(G.north);

\end{tikzpicture}

%% file: frames/tikzfig_CGAN2.tex
\begin{tikzpicture}[scale=1,
roundnode/.style={circle, draw=orange!80!black, fill=orange!10, very thick, minimum size=30},
squarednode/.style={rectangle, draw=cistibpurple!80, fill=cistibpurple!10, very thick, minimum size=1},line width = 0.2mm,node distance=2cm]

\node at (0, 2)   (z) {$z$};
\node at (0, 1)   (y) {$\bar{y}$};
\node at (0, 0)   (x) {$x$};
\node[roundnode] at (1.5, 1)     (F) {$F$};
\node[roundnode] at (3, 1.5)   (G) {$G$};
\node[roundnode] at (6, 0.5)   (D) {$D$};
\node at (4.5, 1.5)   (x2) {$x$};
\node at (7.5, 0.5)   (p) {$p$};

\draw[->] (z)--(G);
\draw[->] (F)--(G);
\draw[->] (F)--(D);
\draw[->] (y)--(F);
\draw[->] (G)--(x2);
\draw[->] (x2)--(D);
\draw[->] (x)--(D);
\draw[->] (D)--(p);
\draw[dashed] (p.north).. controls +(up:1.5) and +(up:0.5) ..(G.north);

\end{tikzpicture}